\documentclass[runningheads]{llncs}
\usepackage{color}
\usepackage{graphicx}
\usepackage{hyperref}
\usepackage{amsmath}   
\usepackage{amssymb}   
\usepackage{booktabs}  
\usepackage{booktabs} 
\usepackage{tabularx}  
\usepackage{multirow}  
\usepackage{marvosym}
\usepackage{float}
\begin{document}
% 标题
\title{Q-Doc: Benchmarking Document Image Quality Assessment Capabilities in Multi-modal Large Language Models}
\titlerunning{Q-Doc}
% 作者

\author{Jiaxi Huang\inst{1} \and
	Dongxu Wu\inst{1} \and
	Hanwei Zhu\inst{2}\textsuperscript{(\Letter)} \and
	Lingyu Zhu\inst{3} \and
	Jun Xing\inst{4} \and
	Xu Wang\inst{5} \and
	Baoliang Chen\inst{1}\textsuperscript{(\Letter)}}

\authorrunning{Huang et al.}
% 机构
\institute{South China Normal University, Guangzhou, China\\
	\email{blchen6-c@my.cityu.edu.hk} \and
	Nanyang Technological University, Singapore  \\
	\email{hanwei.zhu@ntu.edu.sg} \and
	City University of Hong Kong, Hong Kong SAR, China \and
	Shenzhen Academy of Inspection and Quarantine, Shenzhen, China \and
	Shenzhen University, Shenzhen, China}
\maketitle              % typeset the header of the contribution

\renewcommand{\thefootnote}{}
\footnotetext{\textsuperscript{(\Letter)}Corresponding author}

% 摘要
\begin{abstract}
	The rapid advancement of Multi-modal Large Language Models (MLLMs) has expanded their capabilities beyond high-level vision tasks. Nevertheless, their potential for \textbf{Document Image Quality Assessment (DIQA)} remains underexplored. To bridge this gap, we propose \textbf{Q-Doc}, a three-tiered evaluation framework for systematically probing DIQA capabilities of MLLMs at coarse, middle, and fine granularity levels. \textbf{\textit{a)}} At the \textbf{\textit{coarse level}}, we instruct MLLMs to assign quality scores to document images and analyze their correlation with Quality Annotations. \textbf{\textit{b)}} At the \textbf{\textit{middle level}}, we design distortion-type identification tasks including single-choice and multi-choice tests for multi-distortion scenarios. \textbf{\textit{c)}} At the \textbf{\textit{fine level}}, we introduce distortion-severity assessment where MLLMs classify distortion intensity against human-annotated references. Our evaluation demonstrates that while MLLMs possess nascent DIQA abilities, they exhibit critical limitations: inconsistent scoring, distortion misidentification, and severity misjudgment. Significantly, we show that \textbf{Chain-of-Thought (CoT)} prompting substantially enhances performance across all levels. Our work provides a benchmark for DIQA capabilities in MLLMs, revealing pronounced deficiencies in their quality perception and promising pathways for enhancement. The benchmark and code are publicly available at: \url{https://github.com/cydxf/Q-Doc}
	
	\keywords{Document Image Quality Assessment \and Multi-modal Large Language Models \and Multi-Level Evaluation \and Chain-of-Thought Prompting}
\end{abstract}
%
%
% 章节
% 第一部分 引言
\section{Introduction}
The rapid progress of multimodal large language models (MLLMs), such as GPT-4o~\cite{hurst2024gpt} and open-source alternatives like DeepSeek-VL2~\cite{wu2024deepseek}, GLM-4V~\cite{hong2024cogvlm2}, mPLUG-Owl3~\cite{ye2024mplug}, Llama-3.2-Vision~\cite{grattafiori2024llama,meta_llama_3_2_11b_vision_instruct} and Co-Instruct~\cite{wu2024towards} has led to a significant leap in general-purpose Artificial Intelligence systems capable of jointly understanding and reasoning over visual and textual inputs. These models have demonstrated remarkable capabilities across various vision-language tasks, including visual question answering~\cite{antol2015vqa}, image captioning~\cite{chen2015microsoft} and visual grounding~\cite{liu2024grounding}. However, existing benchmarks primarily emphasize high-level semantic understanding~\cite{li2023seed,liu2024ii} or general visual recognition~\cite{liu2024mmbench,fu2024mmecomprehensiveevaluationbenchmark}, leaving the capacity of MLLMs in \textbf{low-level visual quality perception}~\cite{koniq10k,fang2020perceptual}, especially in task-specific domains such as document images, largely underexplored.
In practice, \textbf{document image quality assessment (DIQA)} plays a pivotal role in downstream tasks like OCR, information retrieval, and document understanding~\cite{alaei2023document,li2018cg,van2020assessing,ye2013document}. The ability to perceive quality degradations such as defocus, underexposure, or motion blur distortion is essential for robust real-world document processing, especially in mobile capture scenarios. While recent efforts like Q-Bench~\cite{wu2024qbench} and BLINK~\cite{fu2024blink} conduct preliminary explorations of low‐level perception in natural images, and DocKylin~\cite{zhang2025dockylin} have made strides in document understanding but have not targeted quality assessment, there exists a lack of systematic benchmarking tailored to document image quality, where semantic integrity is tightly coupled with layout, text readability, and distortion sensitivity.

\begin{figure}[htbp]
	\centering
	\includegraphics[width=0.8\linewidth]{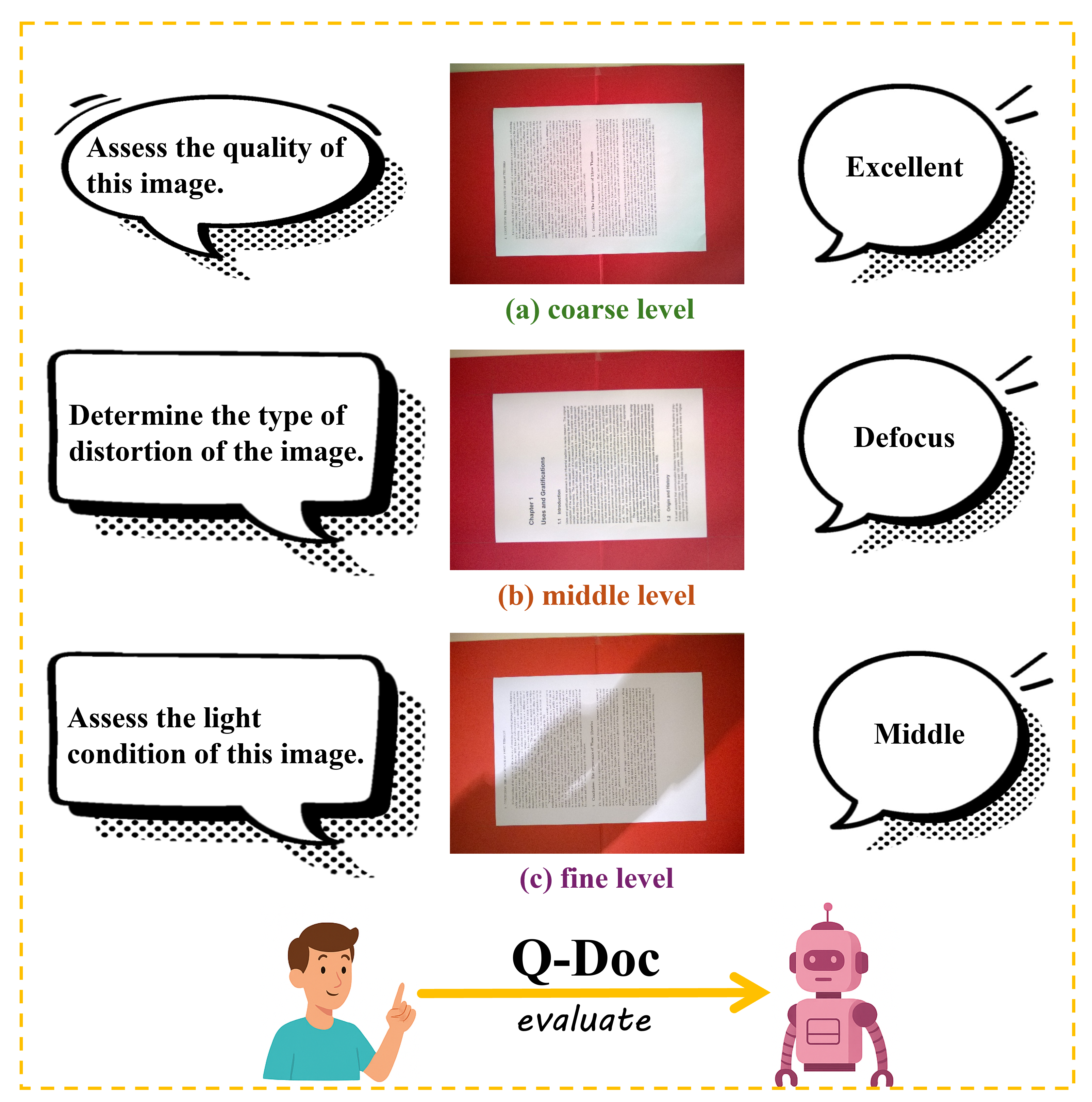}
	\caption{Overview of the proposed Q-Doc benchmark. Q-Doc evaluates MLLMs across three levels: coarse-level quality scoring, middle-level distortion classification, and fine-level severity estimation, using real-world distorted document images.}
	\label{fig:qdoc-overview}
\end{figure}

To bridge this gap, we propose \textbf{Q-Doc} , a comprehensive benchmark for evaluating the DIQA capabilities of MLLMs. Unlike previous works that focus on aesthetic or photographic quality in natural scenes, Q-Doc targets MLLMs' reasoning and judgment abilities on task-oriented, text-centric images captured in real-world mobile settings. Built upon the \textit{SmartDoc-QA} dataset~\cite{nayef2015smartdoc}, Q-Doc includes 4,260 document images with both single and compound distortions across 30 unique documents from three practical domains: modern forms, historical letters, and receipts. As illustrated in Figure~\ref{fig:qdoc-overview}, our benchmark introduces a \textbf{Coarse-Middle-Fine} three-level evaluation framework that progressively dissects the quality perception ability of MLLMs across abstraction levels.
\begin{itemize}
	\item[$\bullet$] \textbf{\textit{Coarse-Level Quality Judgment.}}
	Can the MLLM provide a consistent overall quality rating (e.g., \textit{excellent}, \textit{poor}) for a document image, aligned with downstream OCR performance? We formulate this as a quality classification task and measure correlation with Quality Annotations using SRCC and PLCC.
	\item[$\bullet$] \textbf{\textit{Middle-Level Distortion Recognition.}}
	Can the MLLM accurately identify the \textit{type} of distortion (e.g., blur, lighting, focus loss) in both single-distortion (single-choice) and multi-distortion (multi-choice) settings?
	\item[$\bullet$] \textbf{\textit{Fine-Level Distortion Severity Assessment.}}
	Can the MLLM assess the \textit{severity} of individual distortions (e.g., mild vs. severe blur) using predefined qualitative levels (\textit{good}/\textit{middle}/\textit{poor})?
\end{itemize}

To further enhance MLLMs' reasoning process, we integrate \textbf{Chain-of-Thought (CoT)} prompting into our evaluation pipeline. By encouraging the model to explicitly reflect on text readability and its correlation with specific visual artifacts, CoT significantly improves interpretability and prediction performance across all three levels.
In summary, our contributions are threefold:
\begin{itemize}
	\item[$\bullet$] We introduce \textbf{Q-Doc}, a document-centric benchmark to systematically evaluate MLLMs on low-level visual perception tasks, focusing on the nuanced interplay between document distortions and textual integrity.
	\item[$\bullet$] We design a \textbf{three-level evaluation framework (Coarse-Middle-Fine)} that decomposes the DIQA task into interpretable subtasks, leveraging a real-world mobile document dataset with rich distortion diversity and fine-grained annotations. This framework enables systematic probing of MLLMs' perception capabilities across quality scoring, distortion classification, and severity estimation.
	\item[$\bullet$] We demonstrate that \textbf{CoT prompting} can meaningfully enhance MLLM performance in document image quality assessment, especially for complex multi-distortion cases.
\end{itemize}

Q-Doc establishes a new testing ground for analyzing and advancing MLLMs’ real-world document understanding from a quality-centric perspective, providing insights for both foundational model development and applied document Artificial Intelligence systems.

% 第二部分 实验
\section{Constructing the Q-Doc}

\subsection{General Principles}
\textbf{Focusing on Document-specific Visual Quality of MLLMs.} Unlike general-purpose MLLM benchmarks~\cite{li2023seed,lu2023evaluation}, which evaluate broad multimodal reasoning capabilities, our proposed \textbf{Q-Doc} emphasizes a focused and layered evaluation of how MLLMs perceive and assess visual distortions in document images. Our design adheres to two core principles: (a) emphasizing the perceptual quality and textual readability of real-world document images; (b) structuring evaluation in a hierarchical fashion to progressively dissect MLLM capabilities across \textbf{\textit{perception}}, \textbf{\textit{classification}}, and \textbf{\textit{estimation of distortion severity}}.

\textbf{Layered Quality Evaluation Framework.} We introduce a \textbf{Coarse-Middle-Fine} three-level evaluation strategy to systematically examine different aspects of document image quality perception. This framework is tailored to simulate real-world use cases such as mobile document scanning and OCR applications. Additionally, we incorporate \textbf{CoT} prompting to investigate whether step-by-step textual reasoning improves model performance in fine-grained quality understanding.

\subsection{Dataset Organization}
Our dataset is curated from the \textit{SmartDoc-QA}, a mobile-captured document image set with diverse degradation types. It consists of \textbf{4,260 real-world document images} covering 30 documents in three categories: modern documents, historical administrative letters, and receipts. Among them, 660 are single-distortion images (180 underexposed, 120 motion-blurred, 240 defocused, 120 distortion-free), while the remaining 3,600 combine multiple degradations (e.g., angle, blur, brightness).The precise distribution of distortion types is detailed in Table~\ref{tab:distortion-distribution}
\begin{table}[htbp]
	\centering
	\caption{Distribution of Distortion Types in the Dataset.}
	\label{tab:distortion-distribution}
	{\footnotesize
		\renewcommand{\arraystretch}{1.2}
		\begin{tabular}{@{}lcc@{}}
			\toprule
			\textbf{Distortion Type} & \textbf{Single-Distortion} & \textbf{Multi-Distortion} \\
			\midrule
			Insufficient Brightness & 180 & 600 \\
			Motion Blur & 120 & 300 \\
			Defocus & 240 & 600 \\
			Insufficient Brightness + Motion Blur & - & 600 \\
			Insufficient Brightness + Defocus & - & 1,200 \\
			No Distortion & 120 & 300 \\
			\addlinespace
			\textbf{Total Images} & \textbf{660} & \textbf{3,600} \\
			\bottomrule
		\end{tabular}
	}
	\vspace{0.2cm}
\end{table}
\subsection{Coarse-Level Assessment}
This tier of \textbf{Q-Doc} examines whether MLLMs can accurately evaluate the \textbf{overall visual quality} of a document image.

\paragraph{Task Setup.} For each image, we prompt the MLLM with:
  
\textit{``Assess the quality of this document image and respond with ONLY one of the following words: excellent (highest quality), good (above average quality), fair (average quality), poor (below average quality), bad (lowest quality). Only respond with the single most appropriate word from the list above, with no additional text or explanation.''}

Each of these choices is mapped to a 5--1 scoring scale. The MLLM is expected to select a single term, which is then converted to a numerical quality prediction.

\paragraph{Quantitative Evaluation.} We compare the model-predicted scores with the \textbf{Quality Annotations} of the same images from the SmartDoc-QA dataset. \textbf{Spearman’s rank correlation coefficient (SRCC)} and \textbf{Pearson’s linear correlation coefficient (PLCC)} are calculated to assess the alignment.

\subsection{Middle-Level Distortion Identification}

In the second tier, we assess whether MLLMs can identify the \textbf{type(s) of distortion} present in a document image.

\subsubsection{Single Distortion (Classification Task)}For each single-distortion image, we pose a 4-option multiple-choice question:

\textit{``Please determine the type of distortion in this document image:  
A. Insufficient brightness  
B. Motion blur  
C. Defocus  
D. No Distortion.  
Please answer the option letters directly.''}  

We use this to evaluate the model’s \textbf{distortion-type recognition accuracy} on 660 single-distortion images.

\paragraph{Balanced vs. Unbalanced Accuracy.}  
Since the number of samples per distortion type in the dataset is not uniform, we report two accuracy metrics:

\begin{itemize}
	\item[$\bullet$] \textbf{Unbalanced Accuracy} (\(Acc_{raw}\)): the standard overall accuracy computed directly over the full dataset, without adjusting for class distribution.
	
	\item[$\bullet$] \textbf{Balanced Accuracy} (\(Acc_{bal}\)): an adjusted metric to mitigate class imbalance by assigning higher weight to under-represented distortion types. Specifically, we adopt an inverse-frequency weighting scheme: for each distortion class \(i\), let \(n_i\) be the number of test samples and \(acc_i\) the per-class accuracy. The final score is computed as:
	
	\begin{equation}
		Acc_{bal} = \sum_{i=1}^{C} w_i \cdot acc_i, \quad \text{where} \quad w_i = \frac{1 / n_i}{\sum_{j=1}^{C} 1 / n_j},
	\end{equation}
	
	where \(C\) denotes the total number of distortion categories. This normalized weighting ensures that classes with fewer samples contribute more to the final accuracy, promoting fairer evaluation under class imbalance.
\end{itemize}

\subsubsection{Multiple Distortions (Multi-label Classification Task)} For each multi-distortion image, we prompt:

\textit{``Please determine the type of distortion of this document image (there may be more than one choice):  
A. Insufficient brightness  
B. Motion blur  
C. Defocus  
D. No Distortion.  
Please answer the option letters directly. If it's a multiple-choice question, separate the options with a comma (e.g. A, B).''}

We compute multi-label classification accuracy under a \textbf{strict matching criterion}: A prediction is considered correct only if all and only the true distortion types are selected (i.e., no extra, missing, or incorrect labels). Any deviation, including over-selection, under-selection, or incorrect selection, is counted as an error.

\paragraph{Option Shuffling for Robustness.}  
To prevent any positional bias in model responses, the order of distortion types is randomized across all test samples. For example, \textit{Defocus} may appear as option C in one question but as option A in another. This ensures that the model’s performance reflects true recognition capability rather than memorization of fixed option sequences.

\paragraph{Balanced Accuracy for Multi-label Case.}  
As with the single-label task, we additionally report both unbalanced and balanced accuracy metrics for multi-distortion classification. Here, class-wise accuracy is computed based on exact-match correctness for each unique distortion combination. Balanced accuracy is calculated using the same inverse-frequency weighting as defined above.

\subsection{Fine-Level Distortion Severity Assessment}

At the fine level, we evaluate whether MLLMs can judge the \textbf{severity of each distortion type} in a document image.

\paragraph{Task Setup.} For each distortion type, the images are manually annotated with one of three severity levels: \textit{good, middle, poor}. We prompt the MLLMs with:

\textit{
``Please evaluate the degree of [specific distortion] in this document image. Choose one of the following options:  
A. good B. middle C. poor.  
Answer with the letter only (A, B, or C).''
}

This task is run for each distortion type independently. The model's predictions are compared against human-annotated labels to compute accuracy.

To ensure precise evaluation, we adopt a stratified query strategy. For \textbf{single-distortion images}, we group the samples by distortion type and ask MLLMs to estimate the severity of that particular distortion. For example, given an image annotated with underexposure, we prompt the model to assess the degree of \textit{brightness degradation} only.

For \textbf{multi-distortion images}, each sample contains both brightness-related and blur-related degradation. To disentangle these effects, we prompt the model twice per image: once to assess the severity of brightness degradation, and once to assess the severity of blur. These responses are evaluated separately for each distortion dimension.

It is noteworthy that while the blur category in multi-distortion images may encompass both defocus and motion-related degradations, we intentionally focus the severity assessment on defocus blur. Compared to motion blur, defocus exhibits more stable and perceptually distinct severity gradations, making it a more reliable target for fine-grained quality evaluation. This design choice enables consistent supervision and facilitates clearer performance comparison across models, while avoiding ambiguity introduced by temporally dynamic artifacts.

\subsection{Chain-of-Thought Prompting for Performance Boost}
\begin{figure}[htbp]
	\centering
	\includegraphics[width=0.85\linewidth]{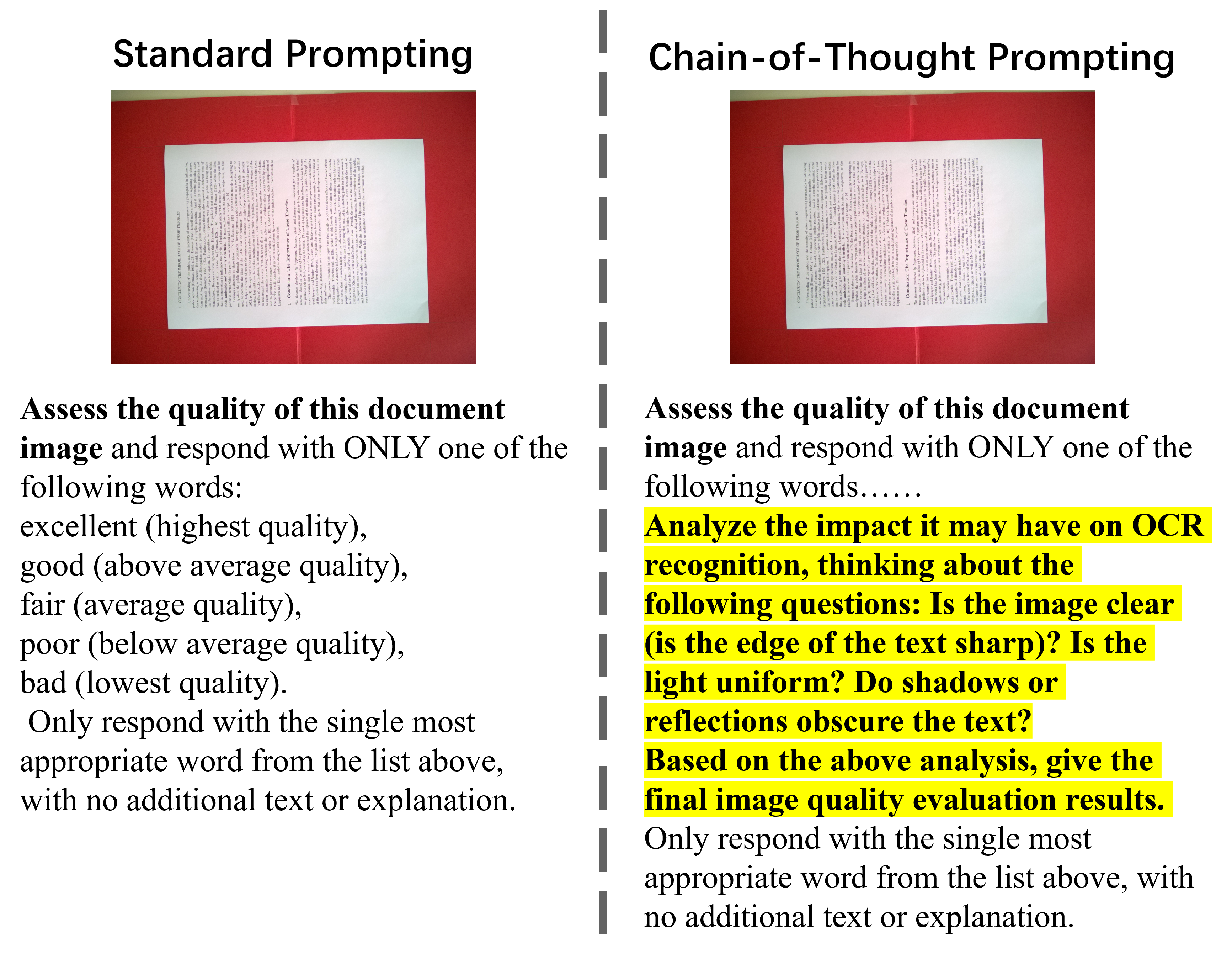}
	\caption{Illustration of Chain-of-Thought (CoT) prompting. The model is encouraged to reason about textual readability and document clarity before producing a quality judgment.}
	\label{fig:cot-prompt}
\end{figure}

To improve interpretability and quality estimation, we introduce \textbf{CoT} prompting strategies, as recent work has shown that CoT significantly enhances performance in MLLMs~\cite{ge2023chain,chen2024visual,zhang2024improve}. For instance, in the coarse-level task, we modify the prompt to:

\textit{``Assess the quality of this document image and respond with ONLY one of the following words: excellent (highest quality), good (above average quality), fair (average quality), poor (below average quality), bad (lowest quality). \textbf{Analyze the impact it may have on OCR recognition, thinking about the following questions: Is the image clear (is the edge of the text sharp)? Is the light uniform? Do shadows or reflections obscure the text? Based on the above analysis, give the final image quality evaluation results.} Only respond with the single most appropriate word from the list above, with no additional text or explanation.''} 
Figure~\ref{fig:cot-prompt} visually highlights how textual clarity and lighting awareness are integrated into step-by-step reasoning.
\paragraph{Ablation Study.} We compare model performance with and without CoT across all three levels. Results show \textit{significant accuracy improvement}, especially at the middle level multi-distortion identification, suggesting that guided reasoning about text readability helps MLLMs better understand visual degradation in documents.

% 第三部分 实验结果
\section{Results on Q-Doc}
\begin{figure}[htbp]
	\centering
	\includegraphics[width=0.9\textwidth]{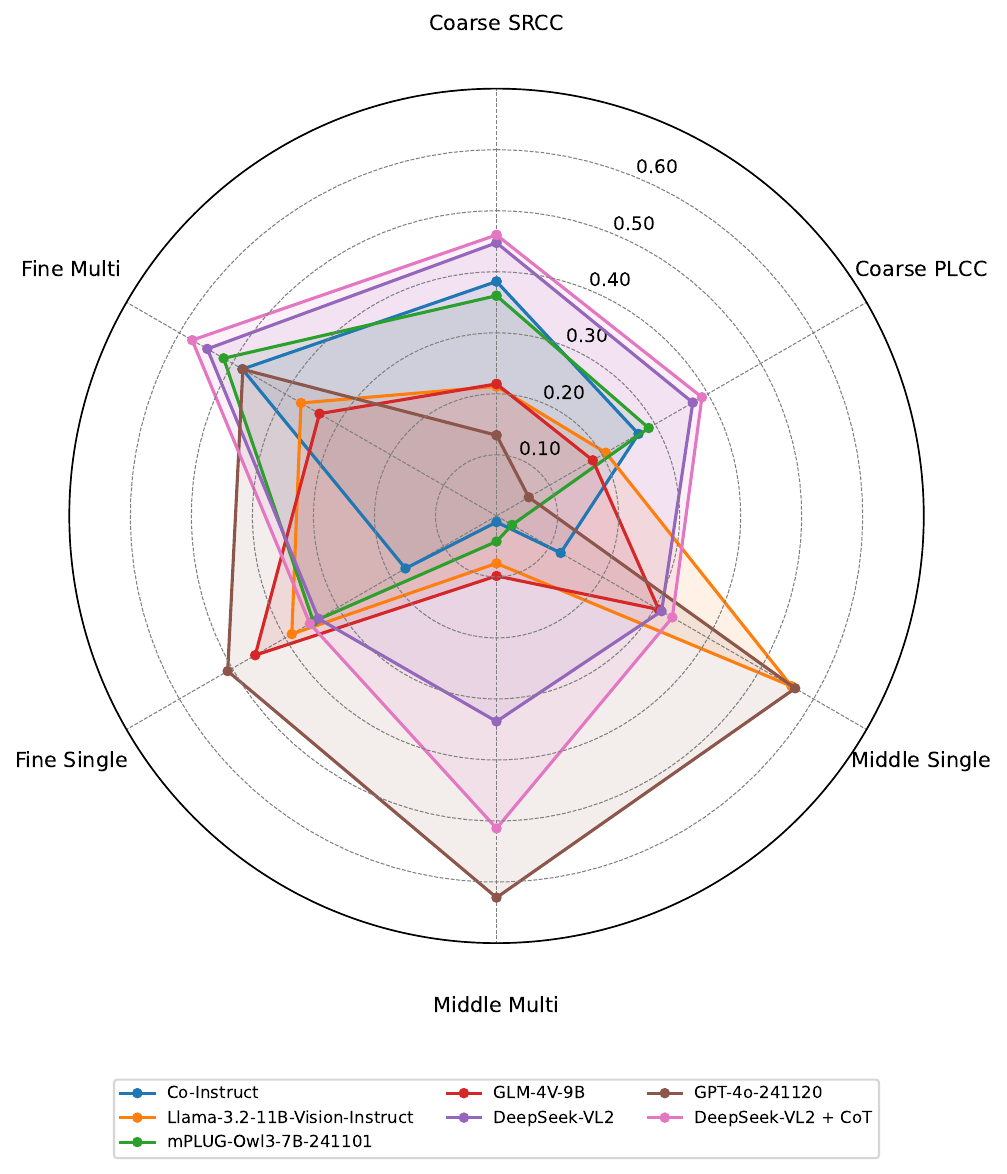}
	\caption{Radar chart of MLLM performance across six quality evaluation metrics: Coarse-level correlation (SRCC, PLCC), Middle-level balanced accuracy (Single/Multiple), and Fine-level balanced accuracy (Single/Multiple).}
	\label{fig:radar}
\end{figure}

We evaluate six representative MLLMs on our proposed three-level Q-Doc. Each model is tested under zero-shot settings, including both competitive open-source MLLMs (Co-Instruct, Llama-3.2-11B-Vision-Instruct~\cite{meta_llama_3_2_11b_vision_instruct}, mPLUG-Owl3-7B-241101, GLM-4V-9B, DeepSeek-VL2) and one leading commercial model (GPT-4o-241120). The experiment follows our Coarse-Middle-Fine framework and is aligned with OCR-grounded quality signals. All tasks are evaluated with both unbalanced and balanced accuracy metrics when applicable.
As shown in Fig.~\ref{fig:radar}, we visualize the overall performance of all MLLMs across six key quality metrics in our benchmark. This provides a quick comparison of model capabilities at each level of evaluation.
\subsection{Coarse-Level: Quality Scoring Correlation with Quality Annotations}

In the Coarse-Level task, we measure how well MLLMs can assess the \textit{overall visual quality} of document images. Each model's output is mapped to a 5-point scale, and its correlation with Quality Annotations is reported using Spearman’s rank correlation coefficient (SRCC) and Pearson’s linear correlation coefficient (PLCC).

\begin{table}[h]
	\caption{Coarse-Level: SRCC and PLCC between MLLM-predicted quality and Quality Annotations. The best results are highlighted in bold, and the second-best results are underlined.}
	\label{tab:coarse}
	\centering
	\begin{tabular}{|
		>{\hspace{5pt}}l<{\hspace{5pt}} 
		|>{\hspace{5pt}}c<{\hspace{5pt}} 
		|>{\hspace{5pt}}c<{\hspace{5pt}} 
		|}
		\hline
		\textbf{Model} & \textbf{SRCC} & \textbf{PLCC} \\
		\hline
		Co-Instruct & 0.3840 & 0.2686 \\
		Llama-3.2-11B-Vision-Instruct & 0.2122 & 0.2069 \\
		mPLUG-Owl3-7B-241101 & 0.3607 & 0.2879 \\
		GLM-4V-9B & 0.2162 & 0.1821 \\
		DeepSeek-VL2 & \underline{0.4474} & \underline{0.3713} \\
		GPT-4o-241120~\textit{(closed-source)} & 0.1321 & 0.0610 \\
		DeepSeek-VL2 + CoT & \textbf{0.4603} & \textbf{0.3886} \\
		\hline
	\end{tabular}
\end{table}

\paragraph{Observations.} Table~\ref{tab:coarse} shows that open-source models such as DeepSeek-VL2 and mPLUG-Owl3 show strong correlation with Quality Annotations, demonstrating their quality-awareness in document perception. Surprisingly, GPT-4o performs significantly worse, despite being a state-of-the-art commercial model.  Notably, DeepSeek-VL2 combined with Chain-of-Thought (CoT) prompting yields a substantial performance gain, highlighting the value of reasoning-focused guidance in scoring document quality.

\subsection{Middle-Level: Distortion Type Recognition}

This level evaluates a model’s ability to identify present distortions, either single (via single-choice) or multiple (via multi-choice), in a document image. To account for label imbalance, we report both unbalanced and balanced accuracy scores.

\begin{table}[h]
	\caption{Middle-Level: Accuracy (\%) on distortion classification. The best results are highlighted in bold, and the second-best results are underlined.}
	\label{tab:middle}
	\centering
	\begin{tabular}{|>{\hspace{5pt}}l<{\hspace{5pt}}|>{\hspace{5pt}}c<{\hspace{5pt}}|>{\hspace{5pt}}c<{\hspace{5pt}}|>{\hspace{5pt}}c<{\hspace{5pt}}|>{\hspace{5pt}}c<{\hspace{5pt}}|}
		\hline
		\multirow{2}{*}{\textbf{Model}} & \multicolumn{2}{c|}{\textbf{Single Distortion}} & \multicolumn{2}{c|}{\textbf{Multiple Distortion}} \\
		\cline{2-5}
		& \(Acc_{raw}\) & \(Acc_{bal}\) & \(Acc_{raw}\) & \(Acc_{bal}\) \\
		\hline
		Co-Instruct & 25.08 & 12.13 & 5.35 & 1.02 \\
		Llama-3.2-11B-Vision-Instruct & 33.74 & \underline{55.80} & 15.57 & 7.78 \\
		mPLUG-Owl3-7B-241101 & 18.54 & 2.95 & 6.00 & 4.20 \\
		GLM-4V-9B & 28.42 & 30.69 & 15.76 & 9.84 \\
		DeepSeek-VL2 & \underline{41.19} & 31.23 & 24.61 & 33.68 \\
		GPT-4o-241120~\textit{(closed-source)} & 38.45 & \textbf{56.50} & \underline{34.54} & \textbf{62.54} \\
		DeepSeek-VL2 + CoT & \textbf{45.74} & 33.26 & \textbf{36.24} & \underline{51.20} \\
		\hline
	\end{tabular}
\end{table}

\paragraph{Observations.} As detailed in Table~\ref{tab:middle}, most MLLMs perform significantly better on single-distortion detection than on the more challenging multi-distortion classification task. GPT-4o achieves the highest balanced accuracy in both categories, while DeepSeek-VL2 combined with CoT demonstrates the most consistent overall performance.

It is worth noting that the accuracy gaps in the multi-distortion task appear larger than in other subtasks. This is primarily due to our \textbf{strict evaluation criterion}: a model’s prediction is considered correct only if it fully matches the ground truth set of distortion types. Any over-selection, under-selection, or incorrect selection leads to a wrong prediction. This design emphasizes the model’s ability to perform precise multi-label classification, but it also makes the task particularly unforgiving, especially when multiple distortions co-occur. Therefore, while the absolute accuracy values may seem low, they still reflect meaningful distinctions in MLLM capability under a high-precision standard.

\subsection{Fine-Level: Severity Estimation}

At this level, we evaluate whether the MLLMs can correctly estimate the severity of visual distortions \textit{(good, middle, poor)}. The task is conducted on both single- and multi-distortion images.

\begin{table}[h]
	\caption{Fine-Level: Accuracy (\%) on distortion severity estimation. The best results are highlighted in bold, and the second-best results are underlined.}
	\label{tab:fine}
	\centering
	\begin{tabular}{|>{\hspace{5pt}}l<{\hspace{5pt}}|>{\hspace{5pt}}c<{\hspace{5pt}}|>{\hspace{5pt}}c<{\hspace{5pt}}|c|}
		\hline
		\textbf{Model} & \multicolumn{2}{c|}{\textbf{Single Distortion}} & \textbf{Multi Distortion} \\
		\cline{2-4}
		& \(Acc_{raw}\) & \(Acc_{bal}\) & Overall Acc. \\
		\hline
		Co-Instruct & 22.62 & 17.23 & 48.08 \\
		Llama-3.2-11B-Vision-Instruct & 25.19 & 38.73 & 36.99 \\
		mPLUG-Owl3-7B-241101 & \underline{55.33} & 34.65 & 51.63 \\
		GLM-4V-9B & 43.68 & \underline{45.65} & 33.49 \\
		DeepSeek-VL2 & 35.14 & 33.68 & \underline{54.71} \\
		GPT-4o-241120~\textit{(closed-source)} & \textbf{68.89} & \textbf{50.86} & 47.99 \\
		DeepSeek-VL2 + CoT & 34.69 & 35.31 & \textbf{57.58} \\
		\hline
	\end{tabular}
\end{table}

\paragraph{Observations.} According to Table~\ref{tab:fine}, GPT-4o achieves the highest accuracy on single-distortion severity estimation, while DeepSeek-VL2 shows the best performance on multi-distortion images. Interestingly, several models, including DeepSeek-VL2 and its CoT-enhanced variant, perform even better on multi-distortion samples than on single-distortion ones.

This result may appear counter-intuitive. However, it stems from differences in task framing rather than evaluation looseness. While both settings evaluate the same two distortion dimensions (brightness and blur) the distribution of distortion subtypes differs across single- and multi-distortion groups. Specifically, our multi-distortion evaluation focuses on well-calibrated and perceptually distinguishable subtypes (e.g., gradual underexposure and controlled defocus), which facilitate more stable assessment conditions. In contrast, single-distortion images include a broader spectrum of sub-variations, potentially increasing intra-class ambiguity.

This design choice reflects our intent to balance task challenge with annotation consistency. By framing multi-distortion severity estimation around more uniformly gradable distortions, we ensure that performance gains reflect model capability rather than dataset variance, thereby reinforcing the diagnostic value of fine-level assessment.
\subsection{Insights on MLLM Performance}

While our evaluation involves six representative MLLMs, our core objective is to understand how these general-purpose models perform across the three hierarchical layers of document image quality perception.

Our findings reveal several important observations:

\begin{itemize}
	\item[$\bullet$] \textbf{Performance inconsistency across levels.} For most models, quality perception ability is \textit{not consistently strong across all levels}. For example, Llama-3.2-11B-Vision-Instruct shows excellent performance in middle-level distortion classification, especially on single distortions, but performs poorly in both coarse-level quality scoring and fine-level severity estimation. In contrast, mPLUG-Owl3-7B-241101 shows decent performance in fine- and coarse-level tasks estimation, yet falls behind significantly in middle-level distortion identification. Even GPT-4o-241120, which excels at both middle-level classification and fine-level granularity, surprisingly underperforms in overall quality scoring, indicating a potential mismatch between high-resolution perception and global document quality judgment.

	\item[$\bullet$] \textbf{DeepSeek-VL2 exhibits the most balanced capability.} Among all models evaluated, DeepSeek-VL2 is the only one that consistently maintains competitive performance across Coarse, Middle, and Fine levels. This balance suggests a more unified and generalizable perception capability, potentially attributable to its \textbf{Mixture-of-Experts (MoE)} architecture. Such architecture may allow the model to dynamically activate specialized sub-modules for different granularities of visual reasoning, enabling both holistic and localized understanding.
	
	\item[$\bullet$] \textbf{Level-specific strengths hint at architectural biases.} The inconsistency of most models across evaluation layers likely stems from architectural or training-phase biases. For example, models like GPT-4o-241120 appear more sensitive to local patterns (reflected in decent fine-level performance) but lack global context alignment (shown in coarse-level correlation drops). These variations suggest that visual token integration, prompt understanding, and training corpus balance may all shape a model’s quality perception profile.
	
	\item[$\bullet$] \textbf{Chain-of-Thought prompting enhances multi-step visual reasoning.} Across all levels, CoT-enhanced DeepSeek-VL2 outperforms its vanilla counterpart, particularly in the Middle and Fine levels. This demonstrates that even for tasks requiring visual-textual alignment, explicit reasoning guidance significantly improves reliability and interpretability of MLLM predictions.
\end{itemize}

In summary, our results indicate that current MLLMs exhibit \textit{fragmented} strengths in document quality assessment, where some models excel in isolated subtasks but lack cross-layer consistency. DeepSeek-VL2, supported by its modular design, stands out as a promising direction toward building quality-aware MLLMs with generalizable and layered perception capabilities.

\section{Conclusion}

In this study, we propose \textbf{Q-Doc}, a comprehensive three-level benchmark designed to evaluate the DIQA abilities of MLLMs. Unlike existing general-purpose benchmarks, \textbf{Q-Doc} explicitly decomposes DIQA into interpretable subcomponents, quality scoring, distortion recognition, and severity estimation, each grounded in real-world document imaging challenges. Our extensive experiments confirm that while today’s MLLMs show promise in OCR-aware perception, they still struggle with distortion interpretation and quantification. We also validate that CoT prompting can significantly improve their diagnostic capabilities. We hope \textbf{Q-Doc} inspires future development of quality-aware, robust, and readable document understanding systems based on MLLMs.
\subsubsection{Acknowledgements} This work was supported in part by the National Natural Science Foundation of China (Grant 62401214), in part by the National Natural Science Foundation of China (Grant 62371310), in part by Shenzhen Science and Technology Program (JCYJ20241202124415021).

\bibliographystyle{splncs04}
\bibliography{refs.bib}
% 参考文献

%\end{thebibliography}
\end{document}